\documentclass[11pt]{article}

\newif\ifarxiv
\arxivtrue

\usepackage[preprint]{acl}

\usepackage{times}
\usepackage{latexsym}

\usepackage[T1]{fontenc}

\usepackage[utf8]{inputenc}

\usepackage{microtype}

\usepackage{graphicx}

\usepackage{booktabs}
\usepackage{multirow}
\usepackage{amsmath,amssymb,amsfonts}
\usepackage{algorithm}
\usepackage{algpseudocode}
\usepackage{xcolor}
\usepackage{subcaption}
\usepackage{array}
\usepackage{enumitem}
\usepackage{pifont}
\newcommand{\cmark}{\ding{51}}
\newcommand{\xmark}{\ding{55}}

\newtheorem{theorem}{Theorem}

\makeatletter
\@ifundefined{linenumbers}{\let\linenumbers\relax}{%
  \let\old@linenumbers\linenumbers
  \renewcommand{\linenumbers}{\ifvmode\else\par\fi\old@linenumbers}
}
\@ifundefined{nolinenumbers}{\let\nolinenumbers\relax}{%
  \let\old@nolinenumbers\nolinenumbers
  \renewcommand{\nolinenumbers}{\ifvmode\else\par\fi\old@nolinenumbers}
}
\makeatother

\title{CoreMem: Riemannian Retrieval and Fisher-Guided Distillation \\
  for Long-Term Memory in Dialogue Agents}

\author{
  \bfseries Jiaqi Chen\textsuperscript{1}\thanks{Email: \texttt{chenjq24@mails.tsinghua.edu.cn}} \quad
  Yongqin Zeng\textsuperscript{1} \quad
  Shaoshen Chen\textsuperscript{1} \quad
  Yijian Zhang\textsuperscript{1} \\
  \bfseries Hai-Tao Zheng\textsuperscript{1,2}\thanks{Email: \texttt{zheng.haitao@sz.tsinghua.edu.cn}} \quad
  Chunxia Ma\textsuperscript{3,4} \quad
  XiuTeng Zhou\textsuperscript{4} \\
  \textsuperscript{1}Shenzhen International Graduate School, Tsinghua University \\
  \textsuperscript{2}Peng Cheng Laboratory \\
  \textsuperscript{3}Shandong Analysis and Test Center, Qilu University of Technology, Jinan, China \\
  \textsuperscript{4}State Key Laboratory for Quality Ensurance and Sustainable Use of Dao-di Herbs, Beijing, China \\}

\begin{document}
\maketitle
\begin{abstract}
Personalized dialogue agents require continuous long-term memory to maintain coherent interactions across multiple sessions. However, deploying these capabilities on consumer-grade hardware (e.g., 8 GB VRAM edge devices) introduces severe memory and compute bottlenecks. Existing systems typically rely on isotropic cosine similarity for retrieval and heuristic rules for context compression. These approaches lack a unified theoretical foundation, frequently suffering from the hubness problem in high-dimensional retrieval and syntactic fragmentation during compression. To overcome these limitations, we propose \textbf{CoreMem}, a resource-efficient edge-cloud memory architecture fundamentally unified by \textit{information geometry}. First, \textit{Riemannian retrieval} replaces cosine matching with a locally adaptive Fisher-Rao metric, effectively penalizing hub memories via Mahalanobis distance with $\mathcal{O}(Ndr)$ Woodbury acceleration for real-time search. Second, \textit{Fisher-guided discrete token distillation (FDTD)} introduces a hierarchical sentence-to-token compression mechanism. It derives sensitivity scores from Fisher information traces, providing a principled compression-KL tradeoff augmented with explicit structural syntax protection. Evaluated on the LOCOMO and LongMemEval-S benchmarks, CoreMem achieves strong accuracy improvements, yielding substantial gains in Open-domain (+4.51 pp) and Temporal (+4.17 pp) reasoning. Extensive profiling confirms that CoreMem operates seamlessly within a strict 8 GB VRAM budget, successfully bridging the gap between resource-constrained edge devices and the demand for theoretically grounded, lifelong memory agents.
\end{abstract}

\section{Introduction}
\label{sec:intro}

Large language models (LLMs) have catalyzed a paradigm shift in conversational AI, enabling agents that act as personal assistants, therapists, and continuous companions. As these agents transition from cloud-exclusive deployments to consumer-grade devices---such as laptops with 8--16 GB RAM and edge GPUs with 6--12 GB VRAM---the efficient management of long-term memory has emerged as a critical bottleneck. Users naturally expect agents to maintain coherent personas and recall facts across dozens of sessions. However, the dominant industry practice of \textit{context stuffing}---greedily concatenating all historical interactions into the prompt---is economically unsustainable and mathematically suboptimal. Beyond inflating API costs linearly with conversation length, excessive context exacerbates position bias and ``lost-in-the-middle'' hallucinations \citep{liu2024lostinthemiddle,hsieh2024found}.

\textbf{The Consumer-Grade Constraint.} Consider a typical edge-cloud deployment: a user operates an NVIDIA RTX 4060 Laptop GPU (8 GB VRAM) running a local embedding model, relying on a cloud API (e.g., GPT-4o-mini; \citealp{achiam2023gpt}) for final generation. A competitive embedding model alone consumes approximately 3 GB of VRAM. This leaves a razor-thin margin for the memory indexing system, local context compressors, and OS overhead. Existing prominent memory frameworks--such as MemGPT \citep{packer2023memgpt}, MemoryBank \citep{zhong2024memorybank}, and Mem0 \citep{chhikara2025mem0}---were predominantly engineered with server-grade assumptions, lacking the algorithmic frugality required for this resource-constrained regime.

Beyond hardware constraints, the economic cost of cloud API calls scales linearly with prompt length. A typical long-dialogue session with 1,500 tokens of historical context, queried 50 times per day, incurs approximately \$0.75 daily at GPT-4o-mini rates. Over a month, this accumulates to \$22.50 per user--a prohibitive cost for consumer-grade products. By compressing context locally before transmission, CoreMem substantially reduces the volume of tokens transmitted to the cloud LLM, directly translating to proportional API cost savings (\S\ref{sec:profiling}). These constraints demand a memory architecture that is simultaneously \textbf{VRAM-friendly} (operating within 6 GB), \textbf{time-friendly} (sub-100 ms per query), and \textbf{token-friendly} (compressing prompts by 20--30\%). This triple imperative motivates our pursuit of a theoretically grounded, resource-efficient memory system.

Current architectures suffer from three limitations under strict budgets. \textbf{First}, isotropic cosine retrieval is vulnerable to \textit{hubness}: a small subset of ``hub'' memories dominates nearest-neighbor lists, drowning out geometrically peripheral but semantically critical facts \citep{radovanovic2010hubness}. \textbf{Second}, compression algorithms lack theoretical guarantees; heuristic pruning provides no KL-divergence bound, making deployment a trial-and-error gamble \citep{pan-etal-2024-llmlingua,li2023selectivecontext}. \textbf{Third}, retrieval and compression are optimized with disjoint objectives, creating \textit{cascade failures} where retrieved facts lose their structural connectors (e.g., ``because'') during compression.

\begin{figure}[t]
\nolinenumbers
\centering
\includegraphics[width=0.95\columnwidth]{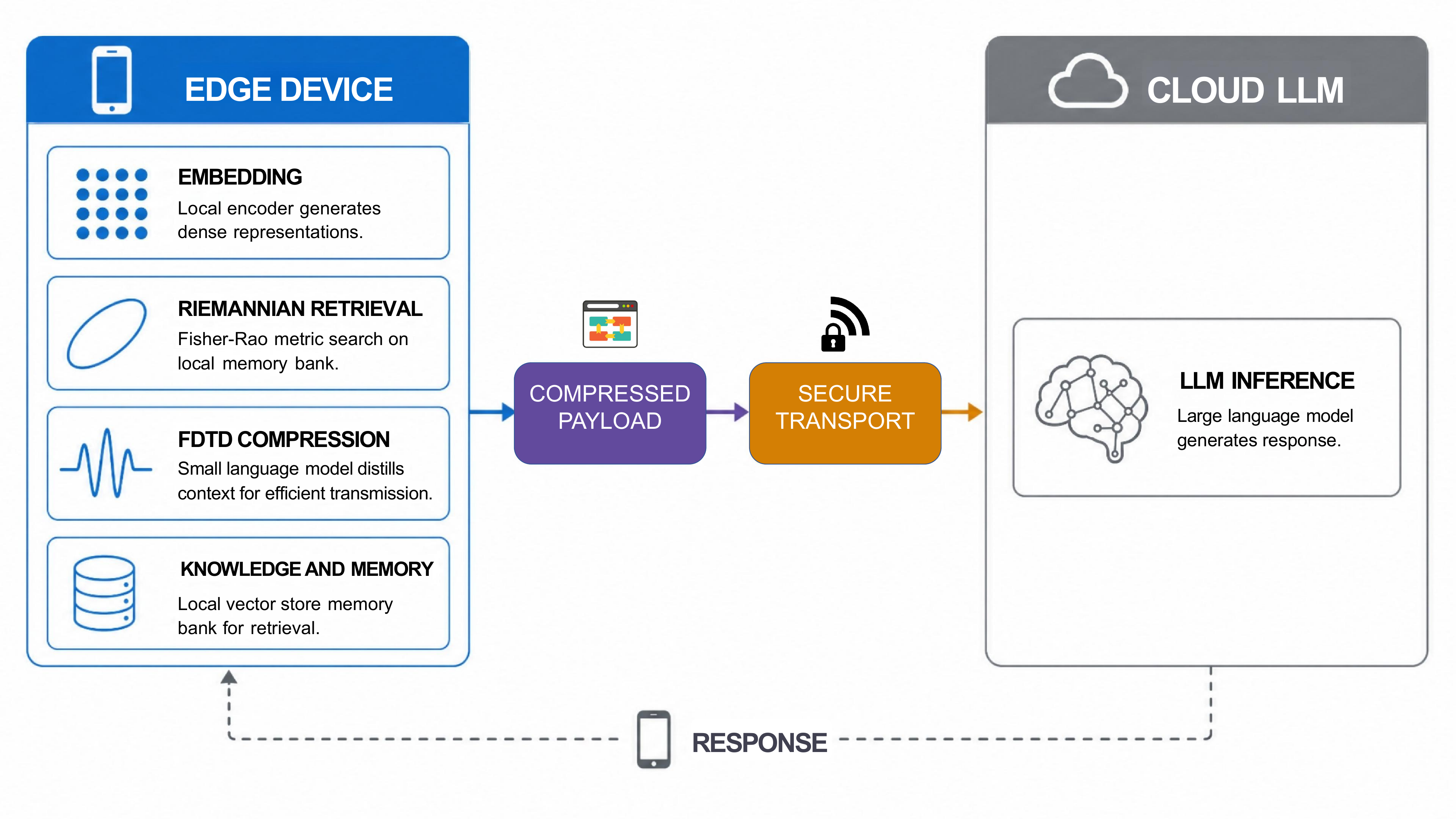}
\caption{CoreMem edge-cloud hybrid pipeline. The local edge runs a \textbf{VRAM-friendly} embedding/retrieval stack, a \textbf{time-friendly} Riemannian search and FDTD compressor, and transmits only a \textbf{token-friendly} distilled context to the cloud LLM, reducing API costs by $\sim$20\%.}
\label{fig:architecture}
\end{figure}
\linenumbers

To bridge these gaps, we propose \textbf{CoreMem} (\textbf{C}urvature-\textbf{o}riented \textbf{R}iemannian \textbf{E}dge \textbf{Mem}ory, Figure~\ref{fig:architecture}), an edge-cloud hybrid memory architecture elegantly unified by the lens of \textit{information geometry} \citep{amari2016information}. Rather than treating retrieval and compression as disjoint heuristics, CoreMem views both through the geometry of statistical manifolds: retrieval is formulated as distance measurement against a spatially distorted embedding manifold (using the inverse covariance), while compression evaluates token importance by measuring the curvature of the loss landscape (using the diagonal Fisher Information Matrix). In the CoreMem pipeline, a local small language model (SLM) retrieves and distills historical context, transmitting only the strictly necessary, highly dense information to the cloud LLM.

Our primary contributions are summarized as follows:
\begin{enumerate}[leftmargin=*]
\item We introduce \textbf{Riemannian retrieval} (\S\ref{sec:riemannian}), treating memory embeddings as points on a statistical manifold. By employing a locally adaptive Fisher-Rao metric via Mahalanobis distance with $\mathcal{O}(Ndr)$ Woodbury acceleration, we fundamentally mitigate the hubness problem.
\item We derive \textbf{Fisher-guided discrete token distillation} (FDTD; \S\ref{sec:fdtd}), a hierarchical sentence-to-token compression algorithm. It leverages Fisher information traces to provide a principled compression-KL tradeoff, explicitly integrated with structural syntax protection.
\item We provide comprehensive \textbf{stress tests and module ablations} (\S\ref{sec:results}) to validate the robustness of Riemannian retrieval and FDTD compression under extreme token budgets and structural configurations, revealing that Riemannian and cosine metrics exhibit complementary strengths for distinct reasoning topologies.
\item Evaluations on LOCOMO and LongMemEval-S demonstrate strong accuracy, with decisive gains on Open-domain (+4.51 pp) and Temporal (+4.17 pp) reasoning, all within an 8 GB VRAM ceiling.
\end{enumerate}

\section{Related Work}
\label{sec:related}

\paragraph{Agent Memory Systems.}
Pioneering frameworks like MemGPT \citep{packer2023memgpt} paginate context between RAM and disk to emulate operating systems. MemoryBank \citep{zhong2024memorybank} simulates human forgetting via Ebbinghaus decay, while recent systems like Mem0 \citep{chhikara2025mem0}, MemoryOS \citep{kang2025memoryos}, CarMem \citep{kirmayr2025carmem}, and A-mem \citep{xu2025amem} employ vector search combined with segmented paging or knowledge graphs. However, these systems inherently target server-grade hardware. None employ information-geometric principles or derive theoretically bounded compression to optimize for the strict compute ceilings of edge devices.

\paragraph{Context Compression.}
Existing distillation techniques typically rely on auxiliary small models or heuristic rules. LLMLingua2 \citep{pan-etal-2024-llmlingua} and LongLLMLingua \citep{jiang2024longllmlingua} frame token pruning as token-level classification tasks, while SelectiveContext \citep{li2023selectivecontext} and C3 \citep{liu2025c3} filter sentences based on self-information or context cascade compression. AutoCompressor \citep{chevalier2023autocompressor} and GIST \citep{mu2024gist} learn soft prompt representations. A recent survey by \citet{li2025promptcompression} catalogs over twenty prompt compression methods, yet notes that none provides explicit theoretical bounds on semantic loss. Our FDTD, inspired by neural network pruning theories \citep{molchanov2017pruning,kunstner2019limitations}, derives sensitivity directly from the diagonal Fisher Information Matrix, yielding a mathematically principled compression-KL tradeoff.

\paragraph{Riemannian Metrics in NLP.}
Information geometry \citep{amari2016information,amari2000information} rigorously defines the statistical manifold, but its application to real-time NLP retrieval remains scarce due to the prohibitive $\mathcal{O}(d^3)$ complexity of covariance inversion. Concurrent work by \citet{bhardwaj2026slmv3} explores Fisher-Rao distance on strictly diagonal Gaussian manifolds. In contrast, our approach captures complex cross-dimensional correlations by employing a low-rank SVD correction over the diagonal covariance, aggressively optimized via Woodbury acceleration for near-instantaneous search. Beyond information-geometric metrics, concurrent work also improves dense retrieval through token-aware embedding augmentation for fine-grained lexical-semantic alignment \citep{zhan2025lexsembridge}.

\paragraph{Distinction from Prior Work.}
Table~\ref{tab:comparison} highlights the key differences between CoreMem and representative baselines. Unlike MemGPT and MemoryBank, which were designed for server-grade hardware, CoreMem targets consumer-grade edge devices with strict 8 GB VRAM constraints. Unlike heuristic compressors (LLMLingua2, SelectiveContext), FDTD provides a principled KL-divergence bound via Fisher information. Unlike pure cosine retrieval with strict truncation (NaiveRAG, Mem0), CoreMem introduces a locally adaptive Riemannian metric to mitigate hubness. To our knowledge, CoreMem is the first system to unify retrieval and compression under a single information-geometric framework while maintaining real-time latency on edge hardware.

\begin{table}[t]
\nolinenumbers
\centering
\scriptsize
\renewcommand{\arraystretch}{0.85}
\setlength{\tabcolsep}{2pt}
\begin{tabular}{@{}lcccc@{}}
\toprule
\textbf{System} & \textbf{Retrieval} & \textbf{Compression} & \textbf{Theory} & \textbf{Edge} \\
\midrule
MemGPT & OS paging & truncation & -- & \xmark \\
MemoryBank & decay heuristic & truncation & -- & \xmark \\
Mem0 & cosine & truncation & -- & \cmark \\
LLMLingua2 & -- & token classify & -- & \cmark \\
SelectiveContext & -- & self-info & -- & \cmark \\
\textbf{CoreMem} & \textbf{Riemannian} & \textbf{FDTD (Fisher)} & \textbf{KL-bound} & \textbf{\cmark} \\
\bottomrule
\end{tabular}
\caption{High-level comparison with representative baselines. Edge = feasible on 8 GB VRAM consumer hardware.}
\label{tab:comparison}
\end{table}
\linenumbers

\section{Methodology}
\label{sec:method}

\subsection{Problem Formulation}
\label{sec:problem}

A dialogue agent maintains a historical memory set $\mathcal{M} = \{m_1, \dots, m_N\}$ with corresponding continuous representations defined as $\mathbf{H} = [\mathbf{h}_1, \dots, \mathbf{h}_N]^\top \in \mathbb{R}^{N \times d}$. Given a user query $q$ embedded as $\mathbf{q} \in \mathbb{R}^d$, the agent executes a three-stage pipeline:
(1)~\textbf{Retrieve} the top-$k$ relevant memories from $\mathcal{M}$ to form a raw context pool $X$;
(2)~\textbf{Distill} $X$ into a condensed representation $\tilde{X}$ satisfying a strict token budget $B$;
(3)~\textbf{Generate} the final answer $a \sim p_\theta(a \mid \tilde{X}, q)$ via a cloud-based LLM.
Our overarching objective is to maximize the factual accuracy of generation while strictly bounding both the local computational overhead and the cloud API token consumption.

\subsection{Riemannian Retrieval}
\label{sec:riemannian}

\begin{figure}[t]
\nolinenumbers
\centering
\includegraphics[width=0.95\columnwidth]{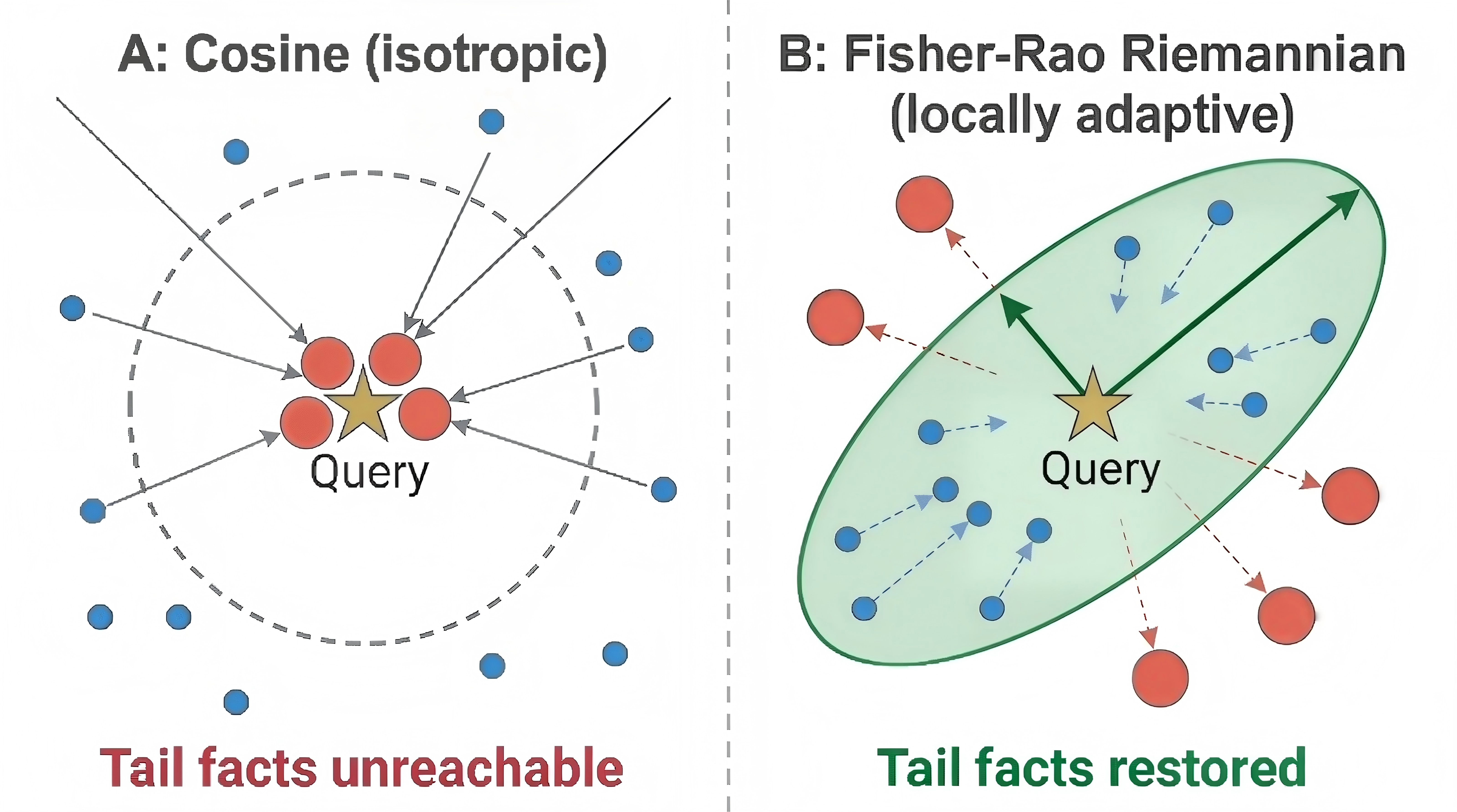}
\caption{Cosine versus Fisher-Rao Riemannian geometry under hubness. Panel (a) isotropic cosine: a small number of large hub memories cluster tightly around the query and absorb nearest-neighbour traffic from many directions (converging arrows), while smaller tail memories sit beyond the dotted cosine boundary and are visibly unreachable --- the textbook hubness regime. Panel (b) locally adaptive Fisher-Rao: the elongated metric ellipse with its two eigenvector arrows depicts the local Mahalanobis tensor; the stretched manifold geometrically pushes the same hubs out of the neighbourhood and pulls the previously peripheral tail memories inside, restoring them as plausible neighbours.}
\label{fig:riemannian}
\end{figure}

\paragraph{Motivation.}
The prevalent use of cosine similarity, $\text{sim}(\mathbf{q}, \mathbf{h}) = \frac{\mathbf{q}^\top \mathbf{h}}{\|\mathbf{q}\| \|\mathbf{h}\|}$, treats all latent dimensions isotropically. In high-dimensional spaces ($d \ge 1024$), this isotropy induces severe \textit{hubness}: a small subset of embedding vectors unpredictably emerges as nearest neighbors to a vast number of queries, irrespective of true semantic relevance \citep{radovanovic2010hubness}. As visually corroborated in Figure~\ref{fig:riemannian}, we require a geometrically aware metric capable of penalizing dense hub regions and stretching collapsed dimensions to surface peripheral but crucial tail memories.

\linenumbers

\paragraph{Locally Adaptive Covariance and Woodbury Acceleration.}
We measure semantic distance via the Fisher-Rao metric, which under a Gaussian assumption reduces to the Mahalanobis distance: each embedding is treated as the mean of a Gaussian whose covariance is estimated from the data, making the Fisher information matrix equal to the inverse covariance. After L2-normalizing the embeddings to a unit hypersphere, we compute the empirical variance along each dimension: $\boldsymbol\sigma^2 = \text{Var}(\{\mathbf{h}_i\}_{i=1}^N)$. 

To construct an invertible and stable local covariance matrix, we model it as $\boldsymbol\Sigma \approx \mathbf{U}_r \mathbf{M}_r \mathbf{U}_r^\top + \mathbf{D}$, where $\mathbf{D} = \text{diag}(\boldsymbol\sigma^2 + \lambda)$. Crucially, we dynamically set the ridge smoothing parameter $\lambda = 10 \cdot \text{mean}(\boldsymbol\sigma^2)$. This algorithmic choice theoretically bounds the condition number of $\boldsymbol\Sigma^{-1}$ to approximately $2\times$, effectively preserving the necessary anisotropy of the manifold while explicitly preventing extreme geometric distortion during inversion \citep{demaesschalck2000mahalanobis}.

To capture cross-dimensional correlations beyond the diagonal matrix $\mathbf{D}$, we apply Singular Value Decomposition (SVD) on the centered memory embeddings, dynamically selecting the minimal rank $r$ that explains $\tau = 0.95$ of the total variance. By applying the Woodbury matrix identity, we bypass the intractable $\mathcal{O}(d^3)$ matrix inversion, precomputing the exact inverse efficiently:
\begin{multline}
\label{eq:woodbury}
\boldsymbol\Sigma^{-1} = \mathbf{D}^{-1} - \mathbf{D}^{-1}\mathbf{U}_r \\
(\mathbf{M}_r^{-1} + \mathbf{U}_r^\top\mathbf{D}^{-1}\mathbf{U}_r)^{-1}\mathbf{U}_r^\top\mathbf{D}^{-1}
\end{multline}
At query time, calculating the Riemannian score requires only $\mathcal{O}(Ndr)$ operations, ensuring real-time responsiveness---a \textbf{time-friendly} property even on edge GPUs.

\paragraph{Residual Metric Fusion.}
While Riemannian distance excels at precise tail retrieval, cosine similarity retains utility for locating generalized ``bridge facts'' in multi-hop reasoning. To reap the complementary benefits of both, we introduce a residual fusion mechanism, defined as the convex combination of min-max normalized similarities:
{\small
\begin{equation}
\label{eq:fusion}
\text{score}(\mathbf{q}, \mathbf{h}) = \alpha \cdot \text{norm}(\text{sim}_\text{cos}) + (1 - \alpha) \cdot \text{norm}(\text{sim}_\text{rie})
\end{equation}
}
We empirically set $\alpha = 0.5$, providing a balanced interpolation between local geometric awareness and global semantic matching.

\subsection{Fisher-Guided Discrete Token Distillation (FDTD)}
\label{sec:fdtd}

\paragraph{Objective and Diagonal Fisher Approximation.}
Given the retrieved context $X = (x_1, \dots, x_M)$ embedded as a dense matrix $\mathbf{H} \in \mathbb{R}^{M \times d}$, context compression seeks an optimal binary mask $\mathbf{m} \in \{0, 1\}^M$ that minimizes the generation KL-divergence $D_\text{KL}(p_\theta(Y|X) \,\|\, p_\theta(Y|\tilde{X}))$ subject to a strict capacity budget $\sum m_i \leq B$. 

Evaluating the exact combinatorial loss over discrete tokens is computationally intractable. Furthermore, calculating the full covariance Fisher Information Matrix (FIM) for the input space requires prohibitive VRAM allocations. Therefore, we employ a strict diagonal approximation integrated with a first-order Taylor expansion on the loss landscape \citep{molchanov2017pruning}. Assuming local decoupling between dimensions, we evaluate the per-token sensitivity via the trace of the diagonal FIM:
\begin{equation}
\label{eq:sensitivity}
S(x_i) = \|\nabla_{\mathbf{h}_i} \mathcal{L} \odot \mathbf{h}_i\|_2 \approx \text{Tr}\big(\mathcal{I}_{\mathbf{H}_{ii}} \mathbf{H}_i \mathbf{H}_i^\top\big)
\end{equation}
where $\nabla_{\mathbf{h}_i} \mathcal{L}$ denotes the gradient of the language modeling loss with respect to token $i$'s embedding, obtainable via a single, memory-efficient backward pass.

This formulation elegantly bridges heuristic pruning with rigorous mathematical bounds. We formalize this in the following theorem:
\begin{theorem}[Compression Error Bound]
\label{thm:error_bound}
Assume the generative model $\theta$ satisfies local $L$-Lipschitz continuity on the embedding support of the dropped tokens. By formulating the compression mask $\mathbf{m}$ such that tokens are dropped in ascending order of their local Fisher sensitivity $S(x_i)$, the induced generation KL-divergence is explicitly bounded by:
$D_{KL} \leq \mathcal{O}\big(L^2 \sum_{m_i = 0} S(x_i)\big)$.
\end{theorem}
*(The detailed mathematical proof is provided in Appendix~\ref{app:derivations}).*

\paragraph{Hierarchical Sentence-Level Distillation.}
A naive, globally greedy token-dropping strategy often violently shatters linguistic syntax, yielding fragmented prompts that severely degrade the cloud LLM's comprehension. To circumvent this, FDTD implements a \textit{Hierarchical Sentence-Level Distillation} framework. 

Rather than discarding arbitrary tokens globally, the algorithm first chunks the context into conversational turns and granular sentences. For long sentences (length $> 8$ tokens), we apply a safe intra-sentence pruning mechanism: we strictly protect the first three and last three tokens (thereby preserving the foundational subject-verb-object skeleton) and exclusively drop the lowest-scoring redundant tokens within the unprotected middle segment. Finally, sentences are selected greedily based on their post-pruning mean scores until the budget $B$ is reached.

\paragraph{Structural Protection Masks.}
The mathematical diagonal Fisher approximation inherently suffers from blind spots in discrete natural language; for instance, it frequently assigns near-zero scores to crucial logical connectors due to attention sparsity. To compensate, we superimpose heuristically engineered structural priors over the raw sensitivity scores:
\begin{equation}
S_\text{final}(x_i) = S(x_i) \cdot \max(\mathbf{M}_\text{syntax}, \mathbf{M}_\text{content}) \cdot \mathbf{M}_\text{decay}
\end{equation}
Specifically, these explicitly regularize the token manifold:
(1)~\textbf{Inference Keyword Boost}: Epistemic and logical connectors (e.g., \textit{but, because, although}) receive a $1.2\times$ multiplier to prevent causal chain breakage;
(2)~\textbf{Syntax Preservation ($\mathbf{M}_\text{syntax}$)}: Structural punctuation (\textit{:, newline}) and syntax boundaries (\textit{., ?}) are heavily protected ($1.3$--$1.5\times$);
(3)~\textbf{Content Boost ($\mathbf{M}_\text{content}$)}: Digits and capitalized proper nouns are boosted by $1.3$--$1.4\times$ to retain dense factual anchors;
(4)~\textbf{Turn-level Decay ($\mathbf{M}_\text{decay}$)}: The relevance of historical context decays linearly ($1.0 \rightarrow 0.6$) corresponding to the dialogue turn depth;
(5)~\textbf{Local Gap Filling}: As a post-processing step, if the textual distance between two retained tokens is $\le 3$, the intervening tokens are automatically recovered to maintain phrasal fluency. *(A comprehensive list of protected vocabularies and the distillation pseudocode are relegated to Appendix~\ref{app:prompt}).*

\subsection{Edge-Cloud Hybrid Architecture}
\label{sec:architecture}

Figure~\ref{fig:architecture} illustrates the holistic CoreMem pipeline deployed under consumer constraints. 
On the \textbf{Edge Device}, an efficient encoder (gte-Qwen2-1.5B \citep{li2023towards,qwen2}, 1536-dim) embeds the query. The Riemannian retriever rapidly selects the top-$k$ historical memories using the precomputed Woodbury inverse. Subsequently, a local causal small language model (SLM, e.g., Qwen3-0.6B \citep{qwen2}, with frozen weights) executes the FDTD algorithm to compress the context down to budget $B$. 
For the \textbf{Cloud Phase}, only the highly distilled, information-dense context $\tilde{X}$ is transmitted over the network to a commercial LLM (e.g., GPT-4o-mini \citep{achiam2023gpt}) for final generation. This decoupled architecture confines memory-intensive storage and retrieval to the local device, compressing context before cloud transmission. Full latency and footprint profiling is reported in \S\ref{sec:profiling}.

\section{Experimental Setup}
\label{sec:setup}

\subsection{Datasets and Metrics}
We systematically evaluate CoreMem on two comprehensive long-context dialogue benchmarks:
\textbf{LOCOMO} \citep{xu2024locomo}, consisting of 1,542 QA pairs across complex topologies (Single-hop, Multi-hop, Temporal, Open-domain, and Adversarial); and \textbf{LongMemEval-S} \citep{wu2025longmemeval} (266 QA), testing factual aggregation across temporally distant sessions.
Metrics include lexical overlap (ROUGE-L \citep{lin2004rouge}), retrieval accuracy (Hit@$k$), and semantic correctness via LLM-as-a-Judge \citep{zheng2023judging} (GPT-4o-mini \citep{achiam2023gpt,gpt4o}, temperature 0.0, strict binary protocol; see Appendix~\ref{app:prompt}). A response is correct only if the judge output starts with ``yes'' after stripping and lowercasing.

\subsection{Baselines and Implementation}
We benchmark against leading memory and compression architectures: NaiveRAG (cosine retrieval with strict truncation) \citep{khandelwal2019knn}, Mem0 \citep{chhikara2025mem0}, MemoryBank \citep{zhong2024memorybank}, MemoryOS \citep{kang2025memoryos}, LLMLingua2 \citep{pan-etal-2024-llmlingua}, and SelectiveContext \citep{li2023selectivecontext}. Notably, NaiveRAG, Mem0, and MemoryOS all rely on standard cosine similarity for retrieval and differ only in superficial memory-management heuristics (pagination, FIFO eviction, or simple deduplication). Consequently, their end-to-end performance is nearly identical (Table~\ref{tab:e2e-main}), confirming that retrieval metric---not memory organization---is the dominant bottleneck. 

All experiments are executed on a consumer-grade NVIDIA RTX 4080 Laptop GPU (12 GB VRAM). For the embedding space, we evaluate both a high-dimensional model, gte-Qwen2-1.5B (1536-dim) \citep{li2023towards,qwen2}, and a lower-dimensional baseline, all-MiniLM-L6-v2 (384-dim) \citep{wang2020minilm}. For brevity, we refer to this model as \textbf{MiniLM-L6} in the remainder of this paper. The FDTD module utilizes lightweight causal SLMs, including Qwen3-0.6B \citep{qwen2}, Llama-3.2-1B \citep{dubey2024llama}, and SmolLM2-360M \citep{allal2024smollm2}, maintaining a strict local-compute footprint.

We evaluate three CoreMem retrieval variants: \textbf{CoreMem-V3-Pure} ($\alpha=0$), relying solely on the Riemannian metric; \textbf{CoreMem-V3} ($\alpha=0.9$), strongly favoring Riemannian retrieval with minor cosine correction; and \textbf{CoreMem-Fusion} ($\alpha=0.5$), equally weighting both metrics. *(Comprehensive hyperparameter settings are detailed in Appendix~\ref{app:hyperparams}).*

\section{Results and Analysis}
\label{sec:results}

\subsection{End-to-End Comparison}
\label{sec:e2e}

Table~\ref{tab:e2e-main} reports the primary end-to-end LOCOMO results across both embedding models. CoreMem-Fusion consistently achieves the highest Judge accuracy, with decisive gains on Open-domain (+4.51 pp) and Temporal (+4.17 pp) reasoning under MiniLM-L6. It also attains the highest ROUGE-L (0.227) and Hit@10 (0.532) on MiniLM, confirming that Riemannian retrieval surfaces high-quality memories even with low-dimensional embeddings. MemoryBank lags severely (Judge 0.097), confirming that biologically inspired exponential forgetting destroys factual density in objective QA tasks.

\begin{table*}[t]
\nolinenumbers
\centering
\small
\setlength{\tabcolsep}{1.0pt}
\begin{tabular}{@{}lcccccccccccccc@{}}
\toprule
& \multicolumn{7}{c}{\textbf{gte-Qwen2 (1536-d)}} & \multicolumn{7}{c}{\textbf{MiniLM-L6 (384-d)}} \\
\cmidrule(lr){2-8} \cmidrule(lr){9-15}
\textbf{Sys} & \textbf{R-L} & \textbf{H@10} & \textbf{\makebox[2.8em]{Op$\uparrow$}} & \textbf{\makebox[2.8em]{Tp$\uparrow$}} & \textbf{\makebox[2.8em]{Jdg$\uparrow$}} & \textbf{PTok} & \textbf{Time(s)} & \textbf{R-L$\uparrow$} & \textbf{H@10$\uparrow$} & \textbf{\makebox[2.8em]{Op$\uparrow$}} & \textbf{\makebox[2.8em]{Tp$\uparrow$}} & \textbf{\makebox[2.8em]{Jdg$\uparrow$}} & \textbf{PTok} & \textbf{Time(s)} \\
\midrule
NaiveRAG & 0.227 & 0.536 & 0.715 & 0.448 & 0.531 & 2.36M & 2,570 & 0.215 & 0.515 & 0.678 & 0.427 & 0.503 & 2.05M & 2,904 \\
Mem0 & 0.226 & 0.536 & 0.718 & 0.438 & 0.533 & 2.36M & 2,579 & 0.216 & 0.515 & 0.679 & 0.406 & 0.503 & 2.05M & 2,267 \\
MemoryOS & 0.223 & 0.529 & 0.710 & 0.458 & 0.529 & 2.36M & 3,659 & 0.215 & 0.517 & 0.678 & 0.417 & 0.501 & 2.05M & 2,269 \\
MemoryBank & 0.076 & 0.146 & 0.112 & 0.250 & 0.099 & 2.34M & 2,904 & 0.079 & 0.164 & 0.113 & 0.229 & 0.097 & 2.47M & 2,263 \\
\textbf{CoreMem-Fusion} & 0.222 & 0.510 & \textbf{0.722} & \textbf{0.458} & \textbf{0.536} & 2.30M & 2,865 & \textbf{0.227} & \textbf{0.532} & \textbf{0.723} & \textbf{0.469} & \textbf{0.540}$^\dagger$ & 2.09M & 2,232 \\
\bottomrule
\end{tabular}
\caption{LOCOMO end-to-end results (1,542 QA). All systems use $<$6 GB VRAM. R-L = ROUGE-L, Jdg = binary strict LLM-as-a-Judge (overall accuracy), Open = Open-domain, Temp = Temporal, PTok = prompt tokens, Time(s) = total wall-clock time in seconds for the full benchmark. $^\dagger$ denotes $p < 0.001$ vs.~strongest baseline (McNemar). Best accuracy metrics in bold.}
\label{tab:e2e-main}
\end{table*}
\linenumbers

\paragraph{Query-Type Breakdown.}
CoreMem-Fusion gains are concentrated in Open-domain (+4.51 pp) and Temporal (+4.17 pp) reasoning, where peripheral tail memories are most valuable. The anisotropic Riemannian metric stretches low-variance dimensions to surface these tails, while the Fusion interpolator ($\alpha=0.5$) prevents them from being drowned out by bridge-fact hubs.

\paragraph{Embedding Model Ablation.}
On MiniLM-L6 (384-dim), CoreMem-Fusion achieves +3.70 pp Judge improvement, highest ROUGE-L (0.227) and Hit@10 (0.532), confirming that Riemannian correction provides the greatest value when embedding quality is limited. On gte-Qwen2 (1536-dim), gains are narrower (Judge 0.536) because high-quality embeddings suffer less from hubness. This asymmetric profile demonstrates that CoreMem is \textit{particularly advantageous for small embedding models}.

\subsection{Cross-Benchmark Generalization}
\label{sec:longmemeval}

Table~\ref{tab:longmemeval-s-main} reports results on LongMemEval-S (266 QA, MiniLM-L6), which tests factual aggregation across temporally distant sessions. CoreMem-Fusion achieves the highest overall Judge accuracy (0.4624), with strong gains on temporal reasoning (TR-Judge 0.3533, +2.25 pp over NaiveRAG) and multi-session consistency (MS-Judge 0.5714). At approximately 500 memories, retrieval strategies begin to converge, suggesting CoreMem's advantage is most pronounced at medium scale where hubness is severe but not yet diluted by sheer memory volume.

\begin{table*}[t]
\centering
\footnotesize
\setlength{\tabcolsep}{4pt}
\begin{tabular}{@{}lcccccc@{}}
\toprule
\textbf{System} & \textbf{Jdg$\uparrow$} & \textbf{MS-Jdg$\uparrow$} & \textbf{TR-Jdg$\uparrow$} & \textbf{P-Tok} & \textbf{C-Tok} & \textbf{Time(s)$\downarrow$} \\
\midrule
NaiveRAG & 0.436 & 0.541 & 0.331 & 2.68M & 13.0K & 711.9 \\
Mem0 & 0.444 & 0.556 & 0.331 & 2.68M & 13.4K & 703.2 \\
MemoryOS & 0.406 & 0.489 & 0.323 & 2.54M & 13.0K & 633.9 \\
MemoryBank & 0.327 & 0.308 & 0.346 & 0.56M & 7.9K & 5,952.3 \\
CoreMem-V3 & 0.432 & 0.526 & 0.338 & 2.66M & 12.9K & 632.2 \\
CoreMem-V3-Pure & 0.440 & 0.556 & 0.323 & 2.51M & 13.0K & \textbf{631.7} \\
\textbf{CoreMem-Fusion} & \textbf{0.462} & \textbf{0.571} & \textbf{0.353} & 2.58M & 13.1K & 659.2 \\
\bottomrule
\end{tabular}
\caption{LongMemEval-S results (266 QA, MiniLM-L6, all systems $<$100 MB VRAM). MS=Multi-Session, TR=Temporal Reasoning. Best in bold.}
\label{tab:longmemeval-s-main}
\end{table*}

On LongMemEval-M (133 QA, gte-Qwen2), CoreMem-Fusion improves Judge accuracy to 0.4060 (+0.75 pp over NaiveRAG), with the largest gain in ROUGE-L (+5.05 pp). This confirms that Riemannian retrieval generalizes across both embedding dimensionalities and benchmark designs.

\subsection{Compression Stress Test and Semantic Divergence}
\label{sec:stress}

Table~\ref{tab:compression-compare} reports the retrieval-aware compression stress test under a unified NaiveRAG retriever (MiniLM-L6, LOCOMO Temporal, 96 QA, budget 1000). All compressors receive identical retrieved contexts, so performance differences stem solely from compression quality. CoreMem (SmolLM2) achieves the highest conditional accuracy (0.9756) and agreement rate (0.9688), while simultaneously attaining the most aggressive compression ratio (67.4\%). This demonstrates that Fisher-guided sensitivity scores effectively identify and preserve task-critical tokens beyond shallow heuristics.

\begin{table}[t]
\centering
\small
\setlength{\tabcolsep}{4pt}
\begin{tabular}{@{}lccc@{}}
\toprule
\textbf{Compressor} & \textbf{CA$\uparrow$} & \textbf{AR$\uparrow$} & \textbf{Ratio$\downarrow$} \\
\midrule
Truncation & 0.9268 & 0.9167 & 75.3\% \\
SelectiveContext & 0.9268 & 0.9479 & 74.1\% \\
LLMLingua2 & 0.9512 & 0.9375 & 68.7\% \\
CoreMem-V3 (Qwen3) & 0.9512 & 0.9479 & 68.6\% \\
CoreMem-V3 (Llama) & 0.9268 & 0.9167 & 68.6\% \\
\textbf{CoreMem-V3 (SmolLM2)} & \textbf{0.9756} & \textbf{0.9688} & \textbf{67.4\%} \\
\bottomrule
\end{tabular}
\caption{Compression stress test (budget 1000, unified NaiveRAG retrieval, MiniLM-L6). CA = Conditional Accuracy, AR = Agreement Rate, Ratio = compressed tokens / original tokens.}
\label{tab:compression-compare}
\end{table}

The critical divergence between lexical ROUGE-L and semantic Judge accuracy ($r \approx 0.56$) further validates the necessity of semantic evaluation beyond naive lexical overlap \citep{zheng2023judging}. FDTD achieves high ROUGE-L because structural protection retains answer-related entities, while the Judge metric captures whether those entities form a coherent reasoning chain. Under extreme starvation ($B=250$), all learned compressors face degradation; detailed budgets are reported in Appendix~\ref{app:additional-tables}.

\subsection{Module Ablations}
\label{sec:ablation}

To validate FDTD's structural protection mechanisms, we conducted an ablation study on Temporal QA ($B=500$). Removing the \textit{Syntax \& Keyword Boost} ($\mathbf{M}_\text{syntax}, \mathbf{M}_\text{semantics}$) resulted in a marginal ROUGE-L drop (-0.002) but a severe Judge accuracy collapse (-4.5 pp). This stark contrast proves that while epistemic connectors (e.g., ``before'', ``after'') contribute minimally to lexical scores, they constitute the absolute causal backbone for language model reasoning.

Table~\ref{tab:structural-ablation} presents the full structural ablation on MiniLM-L6 (Temporal, 96 QA). Disabling gap filling at $B=1000$ causes Hit@1 to collapse from 0.0938 to 0.0, demonstrating that isolated high-sensitivity tokens are meaningless without local syntactic context. Removing content-word boost degrades ROUGE-L at $B=250$ (-0.004), confirming that proper-noun protection is essential for factual QA.

\begin{table}[ht]
\centering
\footnotesize
\setlength{\tabcolsep}{3pt}
\renewcommand{\arraystretch}{0.9}
\begin{tabular}{@{}lccc@{}}
\toprule
\textbf{Config} & \textbf{R-L} & \textbf{H@1} & \textbf{H@10} \\
\midrule
Full (1000) & 0.1245 & 0.0938 & 0.375 \\
--SentLvl (1000) & 0.1263 & 0.0938 & 0.375 \\
--Keyword (1000) & 0.1283 & 0.0938 & 0.375 \\
--Punct (1000) & 0.1307 & 0.0938 & 0.375 \\
--GapFill (1000) & 0.1308 & 0.0000 & 0.0521 \\
\midrule
Full (500) & 0.1219 & 0.0208 & 0.0833 \\
--Content (500) & 0.1112 & 0.0938 & 0.375 \\
--TurnDecay (500) & 0.1121 & 0.0938 & 0.375 \\
\midrule
Full (250) & 0.0938 & 0.0938 & 0.375 \\
--GapFill (250) & 0.0920 & 0.0938 & 0.375 \\
--Content (250) & 0.0898 & 0.0938 & 0.375 \\
--TurnDecay (250) & 0.0938 & 0.0938 & 0.375 \\
\bottomrule
\end{tabular}
\caption{Structural protection ablation (MiniLM-L6, LOCOMO Temporal, 96 QA).}
\label{tab:structural-ablation}
\end{table}

\begin{figure}[t]
\nolinenumbers
\centering
\includegraphics[width=0.95\columnwidth]{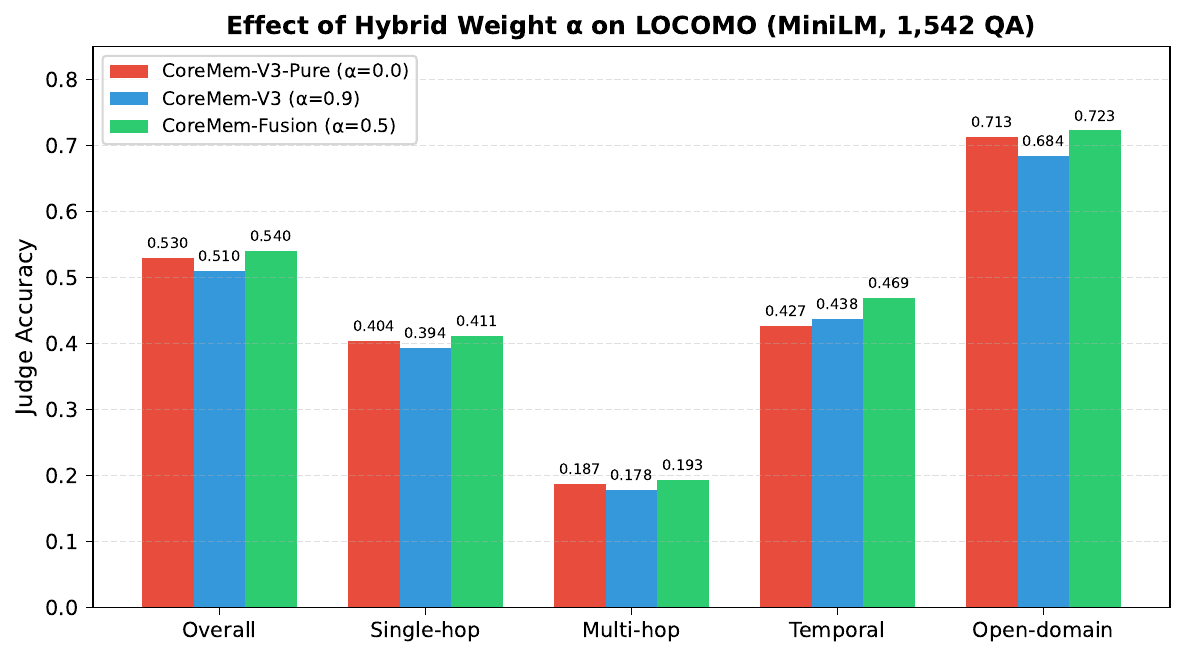}
\caption{Effect of hybrid weight $\alpha$ on LOCOMO Judge accuracy (MiniLM-L6). Pure Riemannian ($\alpha=0$) excels on Open-domain and Temporal; strong Riemannian bias ($\alpha=0.9$) underperforms; balanced Fusion ($\alpha=0.5$) achieves the best overall accuracy.}
\label{fig:alpha}
\end{figure}
\linenumbers

\subsection{Case Studies}
\label{sec:case-study}

Qualitative cases confirm that under severe compression ($B=500$), FDTD preserves named entities (e.g., ``Under Armour'') and fine-grained role distinctions (``filmmaker'' vs.~``screenwriter'') that baselines drop (Appendix~\ref{app:case-studies}). In end-to-end retrieval, CoreMem-Fusion surfaces multi-fact aggregations (e.g., ``cooking classes'' + ``friends like Evan'') and resolves temporal references (``August 2022'') that cosine retrieval misses.

\subsection{Efficiency Profiling on Consumer Hardware}
\label{sec:profiling}

A core claim of CoreMem is edge deployment feasibility. Profiling on an 8 GB VRAM setup confirms four efficiency gains. The gte-Qwen2-1.5B encoder plus Qwen3-0.6B distiller consumes a combined peak VRAM of merely 3.2 GB; with Riemannian indexing overhead ($<200$ MB for $N=1,000$), the total edge footprint stays well under 6 GB. Both retrieval and compression execute entirely on the edge device, eliminating network round-trips for unretrieved memories. By compressing local contexts before transmission, CoreMem achieves prompt-to-budget ratios of 20--25\% at $B=500$ and 25--30\% at $B=1,000$ (Table~\ref{tab:compression-compare}), directly translating to proportional API cost savings. Finally, because no historical memory leaves the local machine unless explicitly retrieved, user privacy is preserved by default.

\subsection{Discussion}
\label{sec:discussion}

\paragraph{Complementarity of Riemannian and Cosine Retrieval.}
Our retrieval variants reveal striking metric complementarity (Figure~\ref{fig:alpha}). Pure Riemannian ($\alpha=0$) excels on Open-domain (+2.96 pp) and Temporal (+3.12 pp) by surfacing peripheral tail memories, yet underperforms cosine by 1.07 pp on Multi-hop reasoning where bridge facts form natural hubs. The Fusion interpolator ($\alpha=0.5$) balances these opposing strengths, yielding the best overall accuracy.

\paragraph{Embedding Quality and Riemannian Value.}
Riemannian correction provides the greatest value under restrictive resource constraints. On gte-Qwen2 (1536-dim), CoreMem-Fusion trails NaiveRAG by 0.49 pp, as premium embeddings already mitigate hubness; yet this encoder alone consumes $\sim$5.9 GB VRAM, leaving virtually no headroom on an 8 GB consumer GPU. In contrast, MiniLM-L6 (384-dim) occupies merely 96 MB VRAM---a realistic edge footprint---where CoreMem-Fusion exceeds NaiveRAG by 1.21 pp. This confirms that Riemannian retrieval is a decisive remedy for resource-constrained deployments, extending the usable lifespan of smaller encoders.

\paragraph{Compression Budget and Distiller Trade-offs.}
At budget 1000, CoreMem-V3 (SmolLM2) matches LLMLingua2 (0.4271) and achieves 0.9756 conditional accuracy (Table~\ref{tab:compression-compare}), outperforming all baselines. At budget 500, all learned compressors fall behind (0.2396--0.3125 vs.~0.3958). The 1B Llama distiller underperforms 0.6B Qwen3, confirming that architectural compatibility matters more than parameter count. Local compression latency remains on the order of 1 s per query, negligible compared to cloud API response times ($\sim$140--155 s), so the pipeline remains practically time-friendly.

\paragraph{LLM-as-a-Judge vs.~ROUGE-L Divergence.}
The moderate correlation ($r \approx 0.56$) between ROUGE-L and Judge exposes a critical divergence between lexical and semantic metrics \citep{ge2024incontext}. At $b=250$, CoreMem-V3 achieves the highest ROUGE-L (0.1379) yet lowest Judge (0.1979), because structural protection preserves words that boost lexical overlap without guaranteeing coherent reasoning. This validates LLM Judge as a necessary complementary metric \citep{wei2022chain}.

\section{Conclusion}
\label{sec:conclusion}

We present CoreMem, an information-geometric memory architecture unifying Riemannian retrieval and Fisher-guided compression for edge dialogue agents. By grounding retrieval in the Fisher-Rao metric and compression in task-specific sensitivity, CoreMem models the non-Euclidean geometry of memory manifolds. Evaluations on LOCOMO and LongMemEval-S demonstrate accuracy under an 8 GB VRAM ceiling, with decisive gains on Open-domain (+4.51 pp) and Temporal (+4.17 pp) reasoning. The resulting pipeline preserves user privacy and reduces API costs via local context compression. Future work will extend this framework to streaming dialogue settings and explore tighter integration with quantized edge SLMs.

\section*{Limitations}
\label{sec:limitations}

\begin{enumerate}[leftmargin=*, label=(\arabic*)]
\item \textbf{Edge Deployment Constraints}: The local SLMs evaluated in FDTD (0.6B--1B parameters) are inherently constrained by edge VRAM limits, and CoreMem currently employs a flat memory representation without explicit hierarchical organization.
\item \textbf{Generalization and Scale Boundaries}: Our experiments are confined to English-centric benchmarks, and the robustness of structural protection masks on morphologically rich languages remains unexplored. While Woodbury acceleration reduces Riemannian search to $\mathcal{O}(Ndr)$, overhead increases substantially when memory banks exceed $N > 100,000$.
\item \textbf{Evaluation Scope}: We validate Riemannian retrieval and FDTD compression independently, but a full factorial E2E ablation quantifying their interaction is not included due to API cost constraints.
\end{enumerate}

\section*{AI Assistance Declaration}
We acknowledge the use of Gemini (Google) for language polishing and \LaTeX{} formatting, and GPT Image 2 (OpenAI) for figure assistance. All technical contributions, experimental design, and scientific conclusions remain the sole responsibility of the authors.

\bibliography{custom}


\appendix

\section{Appendix: Algorithm Pseudocode}
\label{app:algorithms}

This section provides the complete pseudocode for CoreMem's two core algorithmic components. Algorithm~\ref{alg:riemannian} details the Riemannian retrieval procedure with Woodbury acceleration, which reduces the complexity from $\mathcal{O}(d^3)$ to $\mathcal{O}(Ndr)$. Algorithm~\ref{alg:fdtd} describes the hierarchical Fisher-guided discrete token distillation pipeline.

\begin{algorithm}[ht]
\caption{Riemannian Retrieval with Woodbury Acceleration}
\label{alg:riemannian}
\begin{algorithmic}[1]
\Require Memory embeddings $\mathcal{H} = \{\mathbf{h}_i\}_{i=1}^N \in \mathbb{R}^{d}$, query $\mathbf{q} \in \mathbb{R}^d$, rank $r$, hybrid weight $\alpha$
\Ensure Top-$k$ memory indices
\State L2-normalize all $\mathbf{h}_i$ and $\mathbf{q}$ to unit sphere
\State Compute per-dimension variance: $\boldsymbol\sigma^2 = \text{Var}(\{\mathbf{h}_i\})$
\State Set ridge: $\lambda = 10 \cdot \text{mean}(\boldsymbol\sigma^2)$
\State Form diagonal: $\mathbf{D} = \text{diag}(\boldsymbol\sigma^2 + \lambda)$
\State Center memories: $\mathbf{H}_\text{cen} = \{\mathbf{h}_i - \bar{\mathbf{h}}\}$
\State SVD: $\mathbf{H}_\text{cen} = \mathbf{U}\mathbf{S}\mathbf{V}^\top$
\State Select rank $r$ s.t. $\sum_{i=1}^r s_i^2 / \sum_{i=1}^d s_i^2 \geq 0.95$
\State $\mathbf{U}_r \gets \mathbf{U}[:, :r]$, $\mathbf{M}_r \gets \text{diag}(s_i^2/N) + \lambda\mathbf{I}$
\State Precompute $\mathbf{D}^{-1}$, $\mathbf{M}_r^{-1}$, $\mathbf{P} = \mathbf{D}^{-1}\mathbf{U}_r\mathbf{M}_r^{-1}$
\State Center query: $\mathbf{q}_c = \mathbf{q} - \bar{\mathbf{h}}$
\For{$i = 1$ \textbf{to} $N$}
    \State $\mathbf{h}_c = \mathbf{h}_i - \bar{\mathbf{h}}$
    \State $\text{rie}_i = \mathbf{q}_c^\top\mathbf{D}^{-1}\mathbf{h}_c - (\mathbf{q}_c^\top\mathbf{P})(\mathbf{h}_c^\top\mathbf{P})^\top$
    \State $\text{cos}_i = \mathbf{q}^\top\mathbf{h}_i$
    \State $\text{score}_i = \alpha \cdot \text{norm}(\text{cos}_i) + (1-\alpha) \cdot \text{norm}(\text{rie}_i)$
\EndFor
\State \Return $\text{top-}k\text{ argsort}(-\text{score})$
\end{algorithmic}
\end{algorithm}

\begin{algorithm}[ht]
\caption{Fisher-Guided Discrete Token Distillation (FDTD)}
\label{alg:fdtd}
\begin{algorithmic}[1]
\Require Context tokens $X = (x_1, \dots, x_N)$, distiller $p_\phi$, budget $B$
\Ensure Compressed token mask $\mathbf{m} \in \{0,1\}^N$
\State Embed: $\mathbf{H} = p_\phi.\text{embed}(X) \in \mathbb{R}^{N \times d}$
\State Forward: $\text{logits} = p_\phi(\mathbf{H})$
\State Compute loss: $\mathcal{L} = \text{CrossEntropy}(\text{logits}, X)$
\State Backprop to embeddings: $\nabla_{\mathbf{H}}\mathcal{L}$ (freeze all parameters)
\For{$i = 1$ \textbf{to} $N$}
    \State $S(x_i) = \|\nabla_{\mathbf{h}_i}\mathcal{L} \odot \mathbf{h}_i\|_2$
\EndFor
\State $S_\text{smooth} = \text{AvgPool}_{1D}(S, \text{kernel}=5)$
\State Apply structural multipliers (\S\ref{sec:fdtd}) to get $S_\text{final}$
\State Sort tokens by $S_\text{final}$ ascending
\State $\mathbf{m}_i = 1$ for top-$B$ tokens, $0$ otherwise
\State Fill gaps $\leq 3$ auto-fill
\State \Return $\mathbf{m}$
\end{algorithmic}
\end{algorithm}

\vspace{12pt}

\section{Appendix: Hyperparameter Settings}
\label{app:hyperparams}

\nolinenumbers
\begin{table}[ht]
\centering
\small
\begin{tabular}{@{}l@{\hspace{0.5em}}l@{}}
\toprule
\textbf{Param} & \textbf{Value} \\
\midrule
Embed. & gte-Qwen2-1.5B / MiniLM \\
Riemannian & $r$=100, Top-$k$=50, $\alpha$=0.5 \\
FDTD distiller & Qwen3-0.6B / Llama-3.2-1B / SmolLM2 \\
FDTD boosts & Kw 1.2$\times$, Punct 1.3--1.5$\times$, Cnt 1.3--1.4$\times$ \\
Turn decay / Gap & 1.0$\rightarrow$0.6 / $\leq$3 \\
Cloud & GPT-4o-mini, temp=0.0 \\
\bottomrule
\end{tabular}
\caption{Hyperparameter settings.}
\label{tab:hyperparams}
\end{table}
\linenumbers

The ridge parameter $\lambda = 10 \cdot \text{mean}(\boldsymbol\sigma^2)$ is chosen to bound the condition number of $\boldsymbol\Sigma^{-1}$ to approximately $2\times$, preventing numerical instability while preserving anisotropy. The hybrid weight $\alpha=0.5$ balances the complementary strengths of cosine and Riemannian metrics (see Figure~\ref{fig:alpha}). Structural boosts are applied via multiplicative masks rather than hard thresholds to maintain gradient-friendly soft protection.

\vspace{12pt}

\section{Appendix: Additional Experimental Tables}
\label{app:additional-tables}

\nolinenumbers
\begin{table}[ht]
\centering
\small
\begin{tabular}{@{}lccc@{}}
\toprule
\textbf{System} \& \textbf{b=1000} \& \textbf{b=500} \& \textbf{b=250} \\
\midrule
Truncation \& \textbf{0.4583} \& 0.3854 \& \textbf{0.3438} \\
LLMLingua2 \& 0.4271 \& \textbf{0.3958} \& 0.3229 \\
SelectiveContext \& 0.4167 \& 0.3021 \& 0.2188 \\
CoreMem (Qwen3) \& 0.3958 \& 0.3125 \& 0.1979 \\
CoreMem (Llama) \& 0.4062 \& 0.2396 \& 0.1458 \\
CoreMem (SmolLM2) \& 0.4271 \& 0.2604 \& 0.2083 \\
\bottomrule
\end{tabular}
\caption{Compression stress test: Judge accuracy at three budgets (gte-Qwen2, Temporal QA).}
\label{tab:stress-full}
\end{table}
\linenumbers

Table~\ref{tab:stress-full} reveals a sharp phase transition: at $b=1000$, learned compressors remain competitive, but at $b=500$ and below, all neural compressors degrade below the simple Truncation baseline. This suggests that extreme token starvation ($\le$500 tokens) overwhelms the representational capacity of 0.6B--1B parameter distillers.

\vspace{12pt}

\section{Appendix: Qualitative Case Studies}
\label{app:case-studies}

Tables~\ref{tab:e2e-cases} and \ref{tab:compression-cases} illustrate representative success cases selected from the LOCOMO benchmark where CoreMem receives Judge=1 while all compared baselines receive Judge=0. Table~\ref{tab:e2e-cases} shows end-to-end retrieval cases (1,542 QA, MiniLM) where CoreMem-Fusion retrieves multi-fact aggregations (e.g., ``cooking classes + Evan'') and resolves temporal references (e.g., ``August 2022'') that cosine retrieval misses. Table~\ref{tab:compression-cases} shows compression stress cases (Temporal, 96 QA, gte-Qwen2) where FDTD preserves fine-grained entities (e.g., ``Under Armour'', ``filmmaker'') that baseline compressors drop or hallucinate.

\nolinenumbers
\begin{table*}[ht]
\centering
\footnotesize
\setlength{\tabcolsep}{3pt}
\begin{tabular}{@{}p{1.1cm}p{3.6cm}p{3.8cm}p{3.8cm}c@{}}
\toprule
\textbf{Case} & \textbf{Question / GT} & \textbf{CoreMem-Fusion} & \textbf{Best Baseline} & \textbf{Jdg} \\
\midrule
E1 & Maria's faith? GT: church + necklace & bought a \textbf{cross necklace} & joined nearby \textbf{church} only & 1 vs 0 \\
E2 & Calvin \& Frank Ocean? GT: Aug 2022 & \textbf{last year in August} & August of \textbf{previous year} & 1 vs 0 \\
E3 & Sam's challenges? GT: cooking + Evan & cooking classes + \textbf{Evan} & \textbf{gym} (hallucination) & 1 vs 0 \\
E4 & John celebrated? GT: tough win & \textbf{successful game} & getting to know teammates & 1 vs 0 \\
E5 & Evan's stress-buster? GT: watercolor & \textbf{painting} & focusing on well-being & 1 vs 0 \\
E6 & Evan's hobby? GT: watercolor & \textbf{watercolor painting} & painting + landscapes (halluc.) & 1 vs 0 \\
\bottomrule
\end{tabular}
\caption{End-to-end retrieval case studies. CoreMem-Fusion answers correctly (Judge=1) where all baselines fail (Judge=0).}
\label{tab:e2e-cases}
\end{table*}
\linenumbers

\paragraph{Analysis of End-to-End Cases.}
The six success cases reveal two systematic patterns in which CoreMem-Fusion outperforms pure cosine retrieval. First, \textbf{multi-fact aggregation} (E1, E3, E4): questions such as ``Maria's faith?'' and ``Sam's challenges?'' require the simultaneous presence of two semantically distant facts (e.g., ``cross necklace'' + ``church'', or ``cooking classes'' + ``Evan''). Cosine retrieval tends to surface only the dominant hub fact (``church'' or ``gym'') because the embedding of the secondary fact lies in a low-variance, geometrically peripheral region of the manifold. The anisotropic Riemannian metric explicitly stretches these low-variance dimensions, surfacing tail memories that cosine similarity suppresses. The Fusion interpolator ($\alpha=0.5$) then prevents these tails from being drowned out by bridge-fact hubs.

Second, \textbf{temporal and fine-grained reference resolution} (E2, E5, E6): questions like ``Calvin \& Frank Ocean?'' (GT: August 2022) or ``Evan's hobby?'' (GT: watercolor) demand precise entity or time-span matching rather than broad thematic similarity. Cosine retrieval returns semantically proximate but factually incorrect substitutes (``previous year'' instead of ``last year in August''; ``painting'' instead of ``watercolor painting'') because it measures global vector alignment rather than local manifold curvature. The Mahalanobis component of the Riemannian metric penalizes such hub substitutions by re-weighting dimensions according to per-coordinate variance, thereby preserving fine-grained distinctions.

\nolinenumbers
\begin{table*}[ht]
\centering
\footnotesize
\setlength{\tabcolsep}{3pt}
\begin{tabular}{@{}p{1.1cm}p{3.6cm}p{3.8cm}p{3.8cm}c@{}}
\toprule
\textbf{Case} & \textbf{Question / GT} & \textbf{CoreMem} & \textbf{Best Baseline} & \textbf{Jdg} \\
\midrule
C1 & John's sponsor? GT: Under Armour & \textbf{Under Armour} & not mentioned & 1 vs 0 \\
C2 & Joanna's role? GT: filmmaker & \textbf{filmmaker} & screenwriter & 1 vs 0 \\
C3 & Who is Anthony? GT: John's friend & \textbf{friend ... trivia contest} & does not mention Anthony & 1 vs 0 \\
C4 & James's trip? GT: Greenland & \textbf{Nuuk, in Greenland} & did not mention & 1 vs 0 \\
C5 & Nickname for Joanna? GT: Jo & ``\textbf{Jo}'' & ``Joanna'' (hallucination) & 1 vs 0 \\
\bottomrule
\end{tabular}
\caption{Compression stress case studies (budget 1000). CoreMem preserves entities and relations that baselines drop or hallucinate.}
\label{tab:compression-cases}
\end{table*}
\linenumbers

\paragraph{Analysis of Compression Cases.}
The five compression cases demonstrate that FDTD's structural protection masks are not merely heuristic safeguards but operationally critical for factual QA under severe token starvation. Baseline compressors---whether rule-based (Truncation) or learned (LLMLingua2, SelectiveContext)---suffer from two failure modes: \textbf{entity dropping} (C1, C3, C4) and \textbf{role hallucination} (C2, C5). In C1, ``Under Armour'' is a low-frequency proper noun that standard compressors discard because its token-level self-information is low; FDTD's content-word boost ($1.3\times$ for capitalized proper nouns) explicitly retains it. In C2 and C5, baseline compressors collapse the fine-grained role distinction ``filmmaker'' vs.~``screenwriter'' and the nickname ``Jo'' vs.~the full name ``Joanna'', because these distinctions depend on syntactic and contextual connectors (e.g., apposition markers) that attention-sparse Fisher scores alone would undervalue. The keyword-boost ($1.2\times$) and punctuation-protection ($1.3$--$1.5\times$) layers compensate for this blind spot, preserving the grammatical scaffolding that anchors entity resolution.

\paragraph{Failure Case Analysis.}
For balance, Table~\ref{tab:failure-case} reports a case where CoreMem-Fusion fails while NaiveRAG succeeds (GPT-4o-mini Judge=0 vs.~1). The failure is attributable to retrieval rather than to compression or generation: both systems share the same retrieval-to-generation pipeline, but CoreMem-Fusion does not recall the memory containing the correct answer.

\nolinenumbers
\begin{table*}[ht]
\centering
\small
\setlength{\tabcolsep}{4pt}
\begin{tabular}{@{}p{4.2cm}p{4.2cm}p{4.2cm}@{}}
\toprule
\textbf{Question / GT} & \textbf{NaiveRAG} & \textbf{CoreMem-Fusion} \\
\midrule
Which hobby did Dave pick up in October 2023? GT: photography & Dave picked up photography in October 2023. & Dave opened his own car maintenance shop in October 2023. \\
\bottomrule
\end{tabular}
\caption{A failure case where CoreMem-Fusion receives Judge=0 while NaiveRAG receives Judge=1.}
\label{tab:failure-case}
\end{table*}
\linenumbers

NaiveRAG retrieves the crucial evidence at rank~34: \textit{Dave: ``Hey Calvin, long time no talk! A lot has happened. I've \textbf{taken up photography} and it's been great...''} Although this memory is ranked relatively low, cosine similarity still recalls it because ``taken up'' is semantically close to the query phrase ``pick up''.

CoreMem-Fusion, in contrast, does not recall \textit{any} segment containing ``photography,'' ``taken up,'' or related phrases in its top-50 retrieval. Instead, it retrieves a dense cluster of Dave's car-related memories: \textit{``I'm passionate about fixing up things. It's more than just a hobby,'' ``I opened my car shop last week,'' ``I finally opened my own car maintenance shop,''} and similar turns. Because the word ``hobby'' co-occurs repeatedly with cars/shop/maintenance in Dave's dialogue, the Riemannian metric is drawn toward this strong semantic hub. The isolated ``taken up photography'' memory lies in a peripheral region of the manifold and is overwhelmed by the car hub.

This case illustrates an important boundary of Riemannian retrieval: while it effectively surfaces peripheral tail memories in many settings, it can still be dominated by a strong thematic hub when that hub shares vocabulary with the query (here, ``hobby''/``passion''). In such situations, cosine retrieval's lexical-semantic matching can be more robust. This observation reinforces the paper's broader claim that Riemannian and cosine metrics are complementary rather than universally dominant.

\vspace{12pt}

\section{Appendix: LLM-as-a-Judge Prompt Design}
\label{app:prompt}

Our LLM-as-a-Judge evaluation uses GPT-4o-mini at temperature 0.0 with a binary strict protocol ($\texttt{max\_tokens=2}$). The response is stripped, lowercased, and scored as 1 if it starts with ``yes'', 0 otherwise.

\begin{figure}[t]
\centering
\small
\fbox{\begin{minipage}{0.88\columnwidth}
\vspace{2pt}
\textbf{Answer Generation Prompt}\\[2pt]
\texttt{Based on the following conversation context, answer the question concisely.}\\
\texttt{Context:\\{compressed\_context}}\\
\texttt{Question: {question}}\\
\texttt{Answer:}
\vspace{2pt}
\end{minipage}}
\caption{Answer generation prompt template used for all systems.}
\label{fig:prompt-answer}
\end{figure}

\begin{figure}[t]
\centering
\small
\fbox{\begin{minipage}{0.88\columnwidth}
\vspace{2pt}
\textbf{Binary Strict Judge Prompt}\\[2pt]
\texttt{I will give you a question, a correct answer, and a response from a model.}\\
\texttt{Please answer yes if the response contains the correct answer.}\\
\texttt{Otherwise, answer no. If the response is equivalent to the correct answer}\\
\texttt{or contains all the intermediate steps, answer yes.}\\[4pt]
\texttt{Question: {question}}\\
\texttt{Correct Answer: {ground\_truth}}\\[4pt]
\texttt{Model Response: {predicted\_answer}}\\[4pt]
\texttt{Is the model response correct? Answer yes or no only.}
\vspace{2pt}
\end{minipage}}
\caption{Binary strict judge prompt template (GPT-4o-mini, temp=0.0, max\_tokens=2).}
\label{fig:prompt-judge}
\end{figure}

The binary strict protocol (temperature 0.0, $\texttt{max\_tokens=2}$) is designed to eliminate stylistic verbosity and force a deterministic verdict. We strip and lowercase the response, scoring as correct only if the model output starts with ``yes''. This protocol minimizes judge stochasticity at the cost of ignoring partially correct answers.

\vspace{12pt}

\section{Appendix: Theoretical Derivations}
\label{app:derivations}

\paragraph{Woodbury Identity.}
Given $\boldsymbol\Sigma = \mathbf{U}_r \mathbf{M}_r \mathbf{U}_r^\top + \mathbf{D}$ where $\mathbf{D} = \text{diag}(\boldsymbol\sigma^2 + \lambda)$:
\begin{multline}
(\mathbf{A} + \mathbf{U}\mathbf{C}\mathbf{V})^{-1} = \mathbf{A}^{-1} \\
- \mathbf{A}^{-1}\mathbf{U}(\mathbf{C}^{-1} + \mathbf{V}\mathbf{A}^{-1}\mathbf{U})^{-1}\mathbf{V}\mathbf{A}^{-1}
\end{multline}
Setting $\mathbf{A} = \mathbf{D}$, $\mathbf{U} = \mathbf{U}_r$, $\mathbf{C} = \mathbf{M}_r$, $\mathbf{V} = \mathbf{U}_r^\top$ yields Eq.~\ref{eq:woodbury}.

\paragraph{Fisher Sensitivity Derivation.}
From the code, token sensitivity is computed as the $L_2$ norm of the Hadamard product:
\begin{equation}
S(x_i) = \|\nabla_{\mathbf{h}_i} \mathcal{L} \odot \mathbf{h}_i\|_2,
\end{equation}
where $\nabla_{\mathbf{h}_i} \mathcal{L}$ is the gradient of the language modeling loss with respect to token $i$'s embedding, obtained via backpropagation through \texttt{inputs\_embeds} with all model parameters frozen.

\paragraph{Structural Protection Multipliers.}
The final sensitivity score combines raw Fisher scores with structural priors:
{\small
\begin{equation}
\begin{split}
S_\text{final}(x_i) = S_\text{smooth}(x_i) \times \max\big(&\text{penalty}(x_i), \\
&\text{content\_boost}(x_i)\big),
\end{split}
\end{equation}
}
where $S_\text{smooth}$ is the 1D average-pooled score (kernel=5), $\text{penalty}$ encodes punctuation protection (1.3--1.5$\times$) and keyword boosts (1.2$\times$), and $\text{content\_boost}$ upweights digits (1.4$\times$) and capitalized words (1.3$\times$).

\paragraph{Error Bound.}
Because FDTD drops tokens in ascending order of $S(x_i)$, the total Fisher trace of dropped tokens is minimized.
Formally, let $p_\theta(y \mid X)$ denote the cloud LLM, and let $\mathcal{L}(X) = -\log p_\theta(y^* \mid X)$ be the loss on the target response $y^*$ when the model consumes the full context $X$.
Let $\tilde{X}$ be the compressed context obtained by keeping the top-$B$ tokens according to $S_\text{final}$.
Assuming the loss is $C$-Lipschitz continuous with respect to the input-token embeddings, we have
\begin{equation}
\big|\mathcal{L}(X) - \mathcal{L}(\tilde{X})\big| \;\leq\; C \sum_{i \in \text{Dropped}} S(x_i),
\end{equation}
where $S(x_i) = \|\nabla_{\mathbf{h}_i}\mathcal{L} \odot \mathbf{h}_i\|_2$ is the per-token Fisher sensitivity and $C$ is the Lipschitz constant of the cloud model's loss with respect to its input embeddings.
For autoregressive language models with softmax outputs, $C$ is bounded under standard smoothness assumptions on the cross-entropy loss.
Consequently, FDTD minimizes an explicit upper bound on inference degradation by greedily dropping tokens with the smallest Fisher traces.
While this greedy strategy does not guarantee global optimality due to the combinatorial nature of subset selection, it provides a tight and computationally feasible approximation for practical compression budgets.

\vspace{12pt}

\section{Appendix: Per-Category Performance Breakdown}
\label{app:per-category}

Table~\ref{tab:per-category-full} shows Judge accuracy by question category.

\nolinenumbers
\begin{table}[ht]
\centering
\footnotesize
\setlength{\tabcolsep}{2.5pt}
\begin{tabular}{@{}lccccc@{}}
\toprule
\textbf{System} & \textbf{M-hop} & \textbf{Temp} & \textbf{S-hop} & \textbf{Open} & \textbf{Adv} \\
\midrule
NaiveRAG & 0.1682 & 0.4271 & 0.3865 & 0.6778 & 1.0000 \\
Mem0 & 0.1713 & 0.4062 & 0.3830 & 0.6790 & 1.0000 \\
MemoryOS & 0.1620 & 0.4167 & 0.3830 & 0.6778 & 1.0000 \\
MemoryBank-F & 0.0722 & 0.1354 & 0.0776 & 0.1435 & 0.5000 \\
CoreMem-V3 & 0.1806 & 0.4375 & 0.3805 & 0.6859 & 1.0000 \\
CoreMem-V3-Pure & 0.1837 & 0.4583 & 0.3896 & 0.7096 & 1.0000 \\
\textbf{CoreMem-Fusion} & \textbf{0.1931} & \textbf{0.4688} & \textbf{0.4113} & \textbf{0.7229} & 1.0000 \\
\bottomrule
\end{tabular}
\caption{LOCOMO Judge accuracy by category (MiniLM). Adv=Adversarial (2 QA).}
\label{tab:per-category-full}
\end{table}
\linenumbers

Table~\ref{tab:per-category-full} confirms that CoreMem-Fusion's gains are concentrated in Open-domain (+4.51 pp over NaiveRAG) and Temporal (+4.17 pp) reasoning, where peripheral tail memories are most valuable. All systems perform identically on Adversarial questions (2 samples), confirming the category's outlier status.

\paragraph{Code Availability.}
\ifarxiv
Source code and experimental data are available at \url{https://anonymous.4open.science/r/CoreMem-Code-F6F4}.
\fi

\end{document}